\documentclass[letterpaper, 10 pt, conference]{ieeeconf}  

\IEEEoverridecommandlockouts                              

\usepackage{graphics} 
\usepackage{graphicx}
\usepackage{float}
\usepackage{times} 
\usepackage{amsmath} 
\usepackage{amssymb}  
\usepackage{algorithm}
\usepackage{algpseudocode}
\usepackage{bm}
\usepackage{empheq}
\usepackage[colorinlistoftodos]{todonotes}
\usepackage{balance}
\title{\LARGE \bf
Online Dynamic Trajectory Optimization and Control for a Quadruped Robot
}

\author{Oguzhan Cebe$^{1}$, Carlo Tiseo$^{1}$, Guiyang Xin$^{1}$, Hsiu-chin Lin$^{2}$, Joshua Smith$^{1}$, Michael Mistry$^{1}$
\thanks{*This work has been supported by the following  grants: EPSRC UK RAI Hubs NCNR (EPR02572X/1) and ORCA(EP/R026173/1), THING in the EU Horizon2020 (ICT-2017-1 780883) and YLSY (Turkey Ministry of National Education Scholarship)}
\thanks{Oguzhan Cebe, Carlo Tiseo, Guiyang Xin, Joshua Smith,  Michael Mistry are with School of Informatics, Institute of Perception, Action and Behaviour,
        University of Edinburgh, EH8 9AB 10 Crichton Street, Edinburgh, United Kingdom
        {\tt\small O.Cebe@sms.ed.ac.uk}}%
\thanks{Hsiu-chin Lin is with School of Computer Science, Department of Electrical and Computer Engineering
        McGill University}%
}

\begin{document}

\maketitle
\thispagestyle{empty}
\pagestyle{empty}

\begin{abstract}

Legged robot locomotion requires the planning of stable reference trajectories, especially while traversing uneven terrain. 
The proposed trajectory optimization framework is capable of generating dynamically stable base and footstep trajectories for multiple steps. 
The locomotion task can be defined with contact locations, base motion or both, making the algorithm suitable for multiple scenarios (e.g., presence of moving obstacles). The planner uses a simplified momentum-based task space model for the robot dynamics, allowing computation times that are fast enough for online replanning. This fast planning capability also enables the quadruped to accommodate for drift and environmental changes. The algorithm is tested on simulation and a real robot across multiple scenarios, which includes uneven terrain, stairs and moving obstacles. The results show that the planner is capable of generating stable trajectories in the real robot even when a box of 15 cm height is placed in front of its path at the last moment.

\end{abstract}
\section{INTRODUCTION}
Legged robots can traverse uneven terrains that are not suitable for wheeled robots. However, legged locomotion on uneven terrain is a challenging task, especially as base motion and footsteps are coupled in terms of kinematics and dynamics. Moreover, locomotion tasks are constrained by the contact dynamics such as unilateral force and friction cone constraints. These challenges make the generation of dynamically feasible motion trajectories an important part of the locomotion problem. 

Nonlinear trajectory optimization (NLOP) is an attractive approach for motion planning and control of legged robots. High level tasks and system dynamics can be represented using cost functions and constraints \cite{dai2014whole}. The formulation of the system dynamics and locomotion parameters is particularly important for the performance of the NLOP. While whole-body dynamics models enable joint reference signal generation and better physics consistency \cite{mastalli2019crocoddyl}, a locomotion task can be described in a more straightforward way with reduced dynamic models  \cite{kim2019highly}. The dynamic model can be simplified further and contact forces and angular momentum dynamics can be excluded from the problem \cite{bellicoso2018dynamic}. Contact location, contact timings and motion can be optimized jointly, effectively increasing the search space of the NLOP  \cite{bledt2019implementing}.

Decoupling the problem into two independent problems (contact planning and motion planning) yields a more tractable optimization problem by constraining the search space and reducing the computation time \cite{deits2014footstep}\cite{mastalli2019crocoddyl}. In contrast, solving the coupled problem enables a better exploitation of the robot's dynamical capabilities \cite{winkler2018gait}\cite{dai2014whole}\cite{mastalli2020motion}\cite{mordatch2012discovery}\cite{neunert2018whole}\cite{carius2018trajectory}. Yet, the existing frameworks find limited use in real-world uneven terrain conditions with changing environments. The algorithms in \cite{farshidian2017real} and \cite{neunert2018whole} have only been implemented on flat terrain. In \cite{melon2020reliable}, non-coplanar terrain locomotion is achieved, but the motion was planned offline. These limitations arise from the fact that NLOP based legged locomotion frameworks depend on good warm starts, carefully tuned task specific cost functions and stable and reachable reference states to act as attractors \cite{garcia2019convex}. Due to the intuition and computation time need, satisfying these conditions in online replanning remains a challenge.
\begin{figure}[t]
\centering
\includegraphics[width=0.6\linewidth]{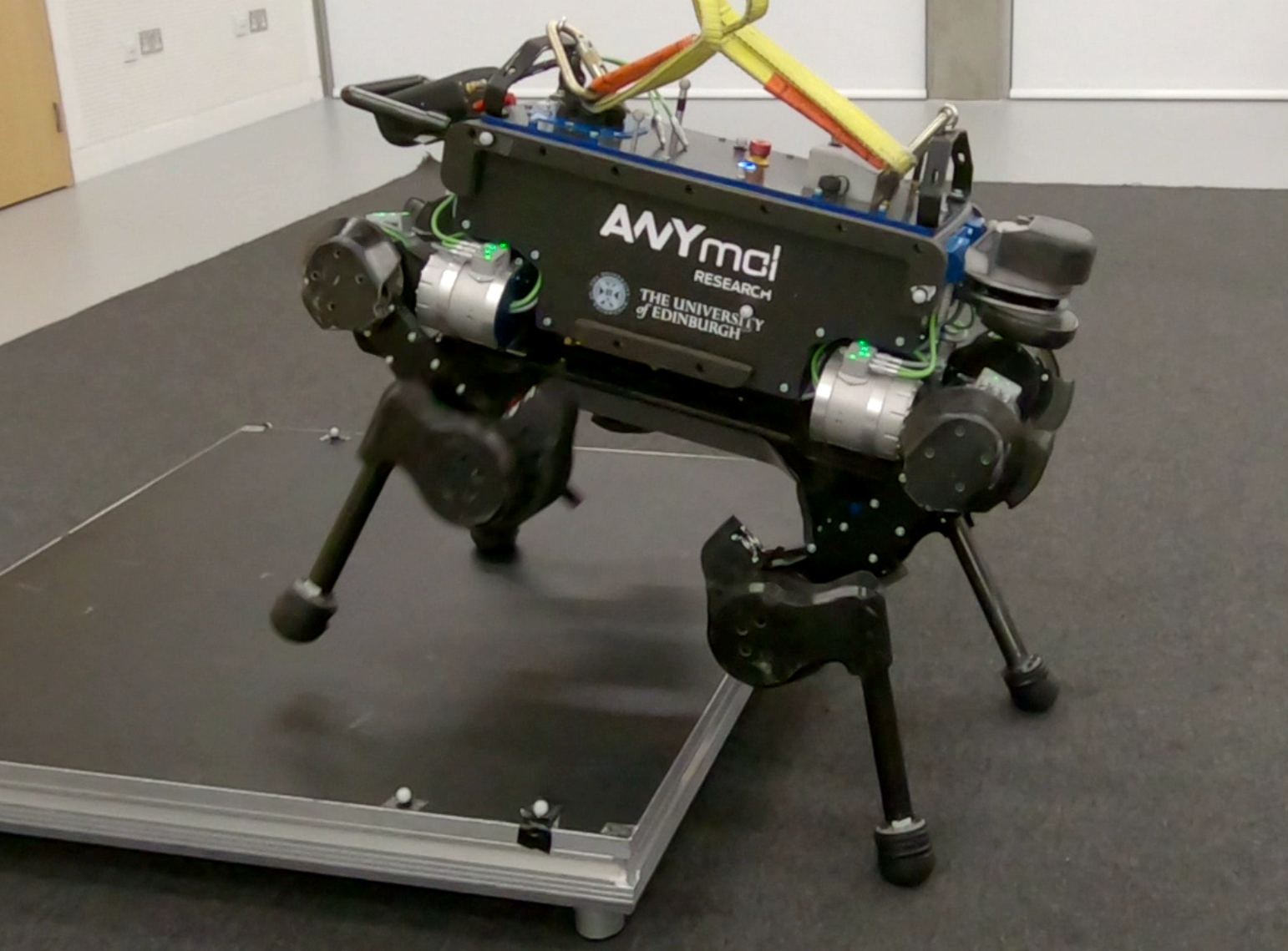}
\caption{A relocated box climbing task is implemented in a real-world experiment. The robot begins a flat terrain locomotion task and a box with 15 cm height is pushed in front of it during motion. The online trajectory optimizer generates a new trajectory that enables the robot to climb on the box.}
\label{fig:real_world}
\end{figure}

Among successful real-world implementations of non-coplanar quadruped locomotion, a 3D mass model with nonlinear zero moment point (ZMP) constraints is used to generate base trajectories, while footsteps are optimized after every base motion optimization \cite{bellicoso2018dynamic}.
In \cite{fankhauser2018robust}, base motion and footsteps are optimized sequentially at every step and a foothold score map is used to find feasible footstep positions. 
In both papers, online footstep optimization increased the reachability, yet individual contact forces were not part of the decision variables and contact stability could not be guaranteed. 
\cite{melon2020reliable} used phase-based end-effector parameterization to generate optimal base, feet and contact timing trajectories with a 16 footstep horizon. The trajectory was not replanned during motion, which limits the capability of adapting to changes in the environment and state drifts. 
In \cite{mastalli2020motion}, both coupled and uncoupled planning were discussed and implemented using a CoP preview model, where coupled planning required significantly longer computation times, but better motion capability.
\begin{figure}[t]
\centering
\includegraphics[width=0.8\linewidth]{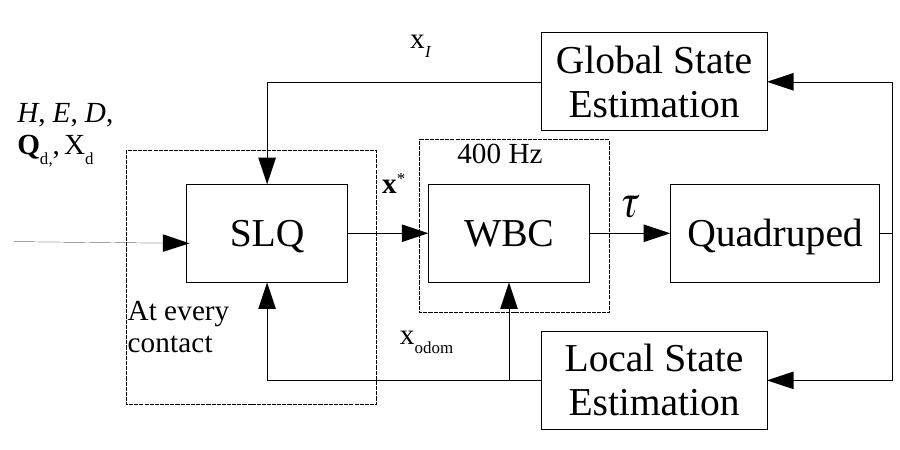}
\caption{An overview of the online trajectory optimization and control framework for quadrupeds. H, E, D are gait horizon length, stance and duration sequences. $\bm{\mathrm{Q}}_{d}$ and $\mathrm{X}_{d}$ are state cost weight and reference. $\mathrm{x}^{*}$ and $\bm{\mathrm{\tau}}$ are state trajectory of SLQ solution and joint torques. $\mathrm{x}_{I}$ and $\mathrm{x}_{odom}$ are state estimations in inertial and odometry frames. There variables are further described in Section \ref{sec:QuadrupedLocomotion}.}
\label{fig_sim}
\end{figure}

There are also methods that do not utilize numerical optimization. In \cite{tiseo2018bipedal}, a generalized inverted pendulum model based on the analysis of the potential energy surface in human locomotion is used to generate non-coplanar biped trajectories and was extended to quadruped trajectory planning in \cite{tiseo2019analytic}. However, these methods do not consider friction cone constraints that are critical to stabilise locomotion upon uneven terrain. Constrained nonlinear trajectory optimization can generate stable trajectories without such assumptions, where the constraints ensure the dynamic feasibility of the problem and cost functions define a high level locomotion task \cite{bledt2019implementing}. Once the decision variables and constraints are selected, this problem can be defined as a Nonlinear Optimal Control Problem (NLOP) to be solved with a NLOP solver.

The proposed trajectory optimization framework removes the restrictions of a decoupled approach, and enables fast online re-planning for dynamic quadruped motion that can cope with terrain alterations, drifts and low frequency disturbances. The planner takes a gait pattern and kinematic tasks as inputs  and returns reference trajectories for base and feet to be passed to the controller. The algorithm works as a standalone planner on convex terrain by discovering feasible base and feet trajectories and achieves collision-free foot placement. On the other hand, for non-convex terrains with contact surface normals provided, a footstep planner or heuristics is required to select the contact surfaces and nominal contact locations for the optimizer. Body and foot plans are passed to a Projected Inverse Dynamics Controller (PIDC) in Task Space (TS-PIDC) \cite{xin2020optimization} for whole-body control (WBC). The impedance behaviour of end-effectors are controlled in the null-space of the contact kinematic constraints, while the contact forces are constrained with higher priority \cite{xin2018model}. 
 Although the trajectories are optimized over a reduced dynamic model, TS-PIDC further guarantees the physical constraints through the use of whole body dynamics while tracking the Cartesian space trajectories.

The proposed planner uses Constrained Sequential Linear Quadratic Programming (SLQ) \cite{giftthaler2018family} with multiple shooting to solve the NLOP. We used the ADRL Control Toolbox (CT) \cite{giftthaler2018control} to encode the nonlinear problem and the High Performance Interior Point Method (HPIPM) \cite{frison2014high}, an efficient solver with inequality constraint support, to solve the constrained linear quadratic sub-problem. 

\section{Quadruped Model}

The reduced momentum model we implement is formulated in task space variables. This formulation reduces the nonlinearity that is introduced by the joint space variables in the system dynamics and its constraints, simplifying the computational complexity of the algorithm and avoiding unneeded calculations.

\subsection{Dynamics Model}
The reduced momentum dynamics model for a single rigid body system with contacts is as follows:
\begin{subequations}
\label{eq:ReducedMomentumModel}
    \begin{align}
    m\ddot{\bm{\mathrm{p}}}&=\sum_{i=1}^{k}\bm{\lambda}_{i}+m\bm{\mathrm{g}}\\
    \bm{\mathrm{I}}_{b}\dot{\bm{\omega}}_{b}+\bm{\omega}_{b}\times \bm{\mathrm{I}}_{b}\bm{\omega}_{b}&=\sum_{i=1}^{k}{(\bm{\mathrm{{r}}_{i}-\bm{\mathrm{p}}}) \times \bm{\lambda}}_{i}\\
    \dot{\bm{\theta}}&=\bm{\mathrm{R}}(\bm{\theta})\bm{\omega}_{b}\\
    {\dot{\bm{\mathrm{r}}}_{i}}&=\bm{\mathrm{{v}}}_{i}
    \end{align}
\end{subequations}
where $m$ is the total mass of the robot, $\bm{\mathrm{g}}$ is the gravitational acceleration, $\bm{\mathrm{I}}_b$ is the torso inertia tensor in base frame, $i$ is the leg index and $k$ is the total number of legs. The states $\bm{\mathrm{p}, \theta}$ are the base position and orientation in XYZ Euler angles, $\dot{\bm{\mathrm{p}}}$, $\bm{\omega}_{b}$ are the base linear velocity and angular velocity in base frame and $\bm{\mathrm{r}}_{i}$ is the position of foot $i$ in base frame as well. Center of mass (CoM) frame is assumed equivalent to base frame and momentum generated by the motion of the legs is neglected. As inputs to the system, we use $\bm{\lambda}_{i}$ as the contact force vector and $\bm{\mathrm{v}}_{i}$ as foot velocity vector. $\bm{\mathrm{R}}(\bm{\theta})$ is the mapping from angular
velocities to Euler angle rates and $\dot{\bm{\theta}}$ is the derivative of the base orientation Euler angles. The state  and control input vectors of the momentum model are as follows,
\begin{align}\label{eq:statesinputs}
\bm{\mathrm{x}}=\begin{bmatrix}
\bm{\mathrm{p}} \; \bm{\theta} \; \dot{\bm{\mathrm{p}}} \; \bm{\omega}_{b} \; \bm{\mathrm{r}}_{i}
\end{bmatrix}^{T}, \quad \bm{\mathrm{u}}=\begin{bmatrix}
\bm{\lambda}_{i} \; \bm{\mathrm{v}}_{i}
\end{bmatrix}^{T}
\end{align}
In \cite{farshidian2017real}, contact forces and joint space velocities have been set as control inputs and the generated trajectories were tracked using a joint space inverse dynamics controller. In our case while both state and control input trajectories are generated with SLQ, only the former in task space is tracked with TS-PIDC. Use of task space variables removes the extra nonlinearity introduced through the joint space variables from the trajectory optimization, as the joint torques are calculated at higher frequency through TS-PIDC for the optimized end effector trajectory, desired task space impedance behaviour and satisfaction of contact force constraints.

\subsection{Constraints}
The input constraints of the dynamic model in the optimization problem are unilateral and linearized friction cone and feet velocity constraints \eqref{eq:subeq1}-\eqref{eq:subeq3} for  active contacts and zero contact force \eqref{eq:subeq4} for inactive contacts. The state constraints are contact position constraint \eqref{eq:subeq5}.
\begin{subequations}
\begin{empheq}[left=\text{stance leg}\Rightarrow\empheqlbrace]{align}
   & \mathrm{b}_{\mathrm{u}, \lambda_{z}}\geq \lambda_{i_{z}}\geq 0, \label{eq:subeq1}\\
    &\bm{\mathrm{U}\lambda}_{i}\leq 0 \label{eq:subeq2},\\
    &\dot{\bm{\mathrm{r}}}_{i}=\bm{\mathrm{0}}, \label{eq:subeq3}\\
    &c_{i}(\bm{\mathrm{r}}_{i}) = \mathrm{0} \label{eq:subeq5}
  \end{empheq}
  \begin{empheq}[left=\text{swing leg}\Rightarrow]{align}
    \bm{\lambda}_{i}=\bm{\mathrm{0}} \quad &\label{eq:subeq4}
  \end{empheq}
\end{subequations}
The function $c_{i}(\bm{\mathrm{r}}_{i})$ represents the contact surface equation. $\mathrm{b_{u}}$ is the upper bound of the contact force box constraint. $\bm{\mathrm{U}}$ is the linearized friction cone inequality constraint.

The unilateral and friction cone constraints in \eqref{eq:subeq1}\eqref{eq:subeq2} enforce contact force stability, whereas end-effector velocity constraint \eqref{eq:subeq3} prevents movements of a foot in stance phase and \eqref{eq:subeq4} ensures no contact force is generated by a swing leg.  \eqref{eq:subeq5} constrains the end effector to lie on a contact surface $c_{i}(\bm{\mathrm{r}}_{i})$. Note that kinematic reachability is not presented as a hard constraint, but as a quadratic cost term in Section \ref{sec:TrajectoryOptimization}.
\section{Quadruped Locomotion}
\label{sec:QuadrupedLocomotion}
The proposed planner assumes that a feasible base and footstep trajectory exist for given contact planes, stance configurations with fixed timings and nominal final state. The contact planes that belong to the stance configurations are used to generate contact surface constraints. Then, the footstep, base sequence and an analytical cost function are used to define the task. The solution of this problem yields dynamically feasible base and feet trajectories that are in the vicinity of the nominal sequences.

The locomotion task is described with a gait pattern, contact planes, running and final cost. By adjusting the final cost weight of base and footsteps, the user can decide which one will be predominant to describe the task. An overview of trajectory optimization framework and controller interface are presented in Fig. \ref{fig_sim} and described in Algorithm \ref{alg:NLOC}.

\subsection{Trajectory Optimization}
\label{sec:TrajectoryOptimization}
In our reduced momentum dynamics model formulation, feet dynamics are modelled in Cartesian space as a first order system to avoid further increasing number of state variables. Continuous swing leg trajectories are generated via a fifth order polynomial spline, with position boundary conditions set by the solution of the optimisation and a desired swing height.

In our formulation, we use quadratic cost functions. This cost formulation has convergence and tuning benefits and its derivatives are immediately known, which reduces the computational complexity \cite{neunert2016fast}. The cost functions are in the form of:
\begin{equation}
\label{eq:RunningCost}
L(t)={}\Tilde{\mathrm{x}}_{d}^{T}\bm{\mathrm{Q}}_{d}\Tilde{\mathrm{x}}_{d}+\Tilde{\mathrm{x}}_{h}^{T}\bm{\mathrm{Q}}_{h}\Tilde{\mathrm{x}}_{h} +\Tilde{\mathrm{u}}^{T}\bm{\mathrm{R}}\Tilde{\mathrm{u}}
\end{equation}
\begin{equation}
\label{eq:FinalCost}
\phi(\mathrm{x}_{t_{f}})={}\Tilde{\mathrm{x}}_{d_{f}}^{T}\bm{\mathrm{Q}}_{d_{f}}\Tilde{\mathrm{x}}_{d_{f}}
\end{equation}
where
\begin{align*}
\Tilde{\mathrm{x}}_{d}={}&\mathrm{x}_{d}-\mathrm{x}(t), \quad \Tilde{\mathrm{x}}_{h}=\mathrm{x}_{h}-\mathrm{x}(t)\\
\Tilde{\mathrm{x}}_{d_{f}}={}&\mathrm{x}_{d_{f}}-\mathrm{x}(t_{f}), \quad \Tilde{\mathrm{u}}=\mathrm{u}(t)\\
\bm{\mathrm{Q}}_{d}=& \text{diag} (\bm{\mathrm{Q}}_{\mathrm{b}}, \bm{\mathrm{Q}}_{\mathrm{fs}}),\quad
\bm{\mathrm{R}}= \text{diag} (\bm{\mathrm{R}}_{\mathrm{\lambda}}, \bm{\mathrm{R}}_{\mathrm{v}})
\end{align*}

The reference states $\mathrm{x}_{d}$ and $\mathrm{x}_{d_{f}}$ represents the desired intermediate and final base pose, twist and footsteps. $\bm{\mathrm{Q}}_{d}$ is the quadratic state cost weight matrix. $\bm{\mathrm{Q}}_{\mathrm{b}}$ and $\bm{\mathrm{Q}}_{\mathrm{fs}}$ are the corresponding weight terms of the base and footstep tasks. The running and final state cost weight matrices $\bm{\mathrm{Q}}_{d}$ and $\bm{\mathrm{Q}}_{d_{f}}$ have the same form, with different weights. This representation provides a straightforward way to describe a high-level task.

Input cost weights $\bm{\mathrm{R}}_{\mathrm{\lambda}}$ and $\bm{\mathrm{R}}_{\mathrm{v}}$ are for the contact force and footstep velocity. In our approach constrained contact forces are used to enforce dynamic feasibility. They are not tracked by the inverse dynamics controller, yet penalizing them provides numerical stability to the optimization. 

We formulate the reachability cost as a linear least-square interpolation between a default stance and a sloped stance, where $\mathrm{x}_{h}$ is the state reference and $\bm{\mathrm{Q}}_{h}$ is the weight matrix (see the Appendix). This suboptimal, but convex approximation reduces the amount of computation by removing their numerical gradient calculations. It should be noted that all quadratic cost weight matrices are diagonal except $\bm{\mathrm{Q}}_{h}$.

By adjusting the state cost weights $\bm{\mathrm{Q}}_{\mathrm{b}}$, $\bm{\mathrm{Q}}_{\mathrm{fs}}$, one can determine if the task is defined predominantly with base pose, twist or footstep positions. With high $\bm{\mathrm{Q}}_{\mathrm{b}}$ low $\bm{\mathrm{Q}}_{\mathrm{fs}}$, the optimizer finds footstep positions that comply with the base task and vice versa. In our implementations, the desired $\bm{\theta}$ and $\bm{\omega_{b}}$ are zero. The resulting orientation state, including the Pitch (Euler Y) angle in Figure \ref{base_p_gap} is due to the reachability cost terms $\bm{\mathrm{Q}}_{h}$ and $\mathrm{x}_{h}$. 
\subsection{Nominal Base Pose and Footstep Position}
Nominal base and footstep sequence $X_{d}$ is a set of points that guide the trajectory optimizer to a feasible solution. $X_{d}$ is encoded to the problem as final cost reference $\bm{\mathrm{Q}}_{d_{f}}$ in Eq. \ref{eq:FinalCost}. We generated $X_{d}$ using a similar approach as in \cite{fankhauser2018robust}, where a geometrical approach with a default stance and footstep length is used to generate nominal footsteps towards the goal. The difference in our method is, we used contact surface planes as constraint \eqref{eq:subeq5} and the nominal footholds as final cost reference to simultaneously optimize base and footstep trajectories, rather than searching the region around the nominal foothold and optimizing footsteps and base pose sequentially. If the nominal foothold is not in contact with the environment, it is projected on the surface plane in vertical direction.

Since we represent the surfaces with their normals instead of scoring the terrain cells for feasibility, our method relies on heuristics to provide clearance margins between the footsteps and non-convex surface edges.

\subsection{Gait Sequence}
\begin{figure}[t]
\centering
\includegraphics[width=0.8\linewidth]{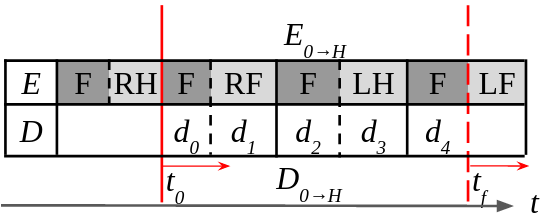}
\caption{Diagram of the quadruped gait sequence for a horizon of $H=5$ (2 footstep) phases. The foot touchdown and lift-off instants are represented as solid and dashed vertical lines on $E$ axis respectively. Light and dark gray $e_{i}$ denotes a swing phase for the corresponding leg ID and dark gray denotes a four feet stance phase. At the touchdown instant $t=t_{0}$ a trajectory of H phases is optimized for $t=t_{0} \rightarrow t_{f}$, where $t_{f}=\sum_{i=0}^{5} d_{i}$. When the new touchdown occurs at $t^{'}_{0}=t_{0}+d_{0}+d_{1}$, a new trajectory is optimized for $t=t^{'}_{0} \rightarrow t^{'}_{f}$, $t^{'}_{f}=\sum_{i=2}^{7} d_{i}$ and the previous trajectory is discarded.}
\label{timing_diagram}
\end{figure}

A gait phase is described by its stance configuration $e_{i} \in$ \{Left Front (LF), Right Front (RF), Left Hind (LH), Right Hind (RH), On Four (F)\} and the corresponding duration $d_{i}$. Three contact configurations are represented with swing leg IDs \{LF,RF,LH,RH\} and four with F. Phase configuration and duration sequences $E$ and $D$ form a gait sequence as seen in Fig. \ref{timing_diagram}. In cyclic gait, each phase is revisited periodically. $E$ and $D$ are determined by using the findings in \cite{fukuoka2015simple} and intuition.

\begin{algorithm}[t]
\caption{NLOP $\And$ Controller Interface}
\label{alg:NLOC}
\begin{algorithmic}[1]
\Statex Stance configuration $e_{i} \in$ \{LF,RF,LH,RH,F\}
\Statex \textbf{Given:}
\Statex For a horizon of $H$ phases
\begin{itemize}
    \item Phase stance sequence $E=[e_{0}, e_{1}, .., e_{H}]$
    \item Phase duration sequence $D=[d_{0}, d_{1}... d_{H}]$
    \item $\mathrm{x}_{0}$ state estimation at touchdown
    \item Desired task $\bm{\mathrm{Q}}_{d}$ and $\mathrm{X}_{d}= [\mathrm{x}_{d_{0}}, \mathrm{x}_{d_{1}}, .., \mathrm{x}_{d_{H}}]$:
\end{itemize}
\Procedure {NLOP}{}
\State $t_{0}=0$,\quad$t_{f}=\sum_{n=j}^{j+H} D_{n} $ 
\State Set final cost state reference $\phi(\mathrm{x}_{t_{f}}) \leftarrow \mathrm{x}_{d_{H}}$
\For {$i \leftarrow 0, H$}
\State Set switched constraints \eqref{eq:subeq1}-\eqref{eq:subeq5} $\leftarrow E_{i}, X_{d_{i}}, D_{i}$
\EndFor
\State Initialize the optimization problem and solve until convergence
\State $\mathrm{x}^{*}(t) \rightarrow \text{TS-PIDC}$ 
\EndProcedure
\end{algorithmic}
\end{algorithm}
\section{Results}
\begin{figure*}[!ht]
\centering
\includegraphics[width=0.8\linewidth]{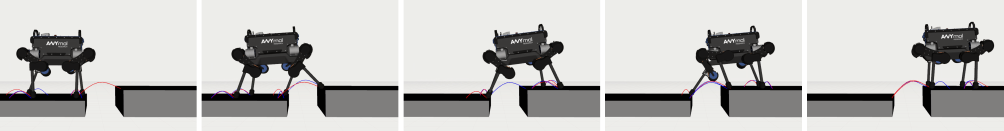}
\caption{Snapshots of the gap crossing simulation. The gap is 30 cm wide and the right box is also 10 cm higher than the left box. Red and blue lines represent the first and second step foot swing plans of the trajectory, respectively.}
\label{fig:snap_sim}
\end{figure*}
An ANYmal quadruped robot \cite{fankhauser2018anymal} has been used for both real experiments and simulations. A VICON motion capture system has been used for measuring the global positions of the robot and the obstacle. The planner and controller run on separate on-board computers and communicate via ROS network. A video of the experimental results can be found in https://bit.ly/3gyccU6

The planner is set with a two footstep horizon on flat terrain forward walking motion; $E=\left[F, RH, F, RF, F \right]$ and $D=\left[0.3, 0.3, 0.25, 0.3, 0.3 \right]$, 1.55 seconds trajectory, with 0.02 seconds sampling time. The constraints \eqref{eq:subeq1}-\eqref{eq:subeq5} are active for their respective phases. Average iteration time for $10^{4}$ tests is 18.340 milliseconds on a consumer grade laptop with 2.80GHz CPU and it takes 4 iterations for convergence without warm-start.

\begin{figure}[thpb]
\centering
\includegraphics[width=0.9\linewidth]{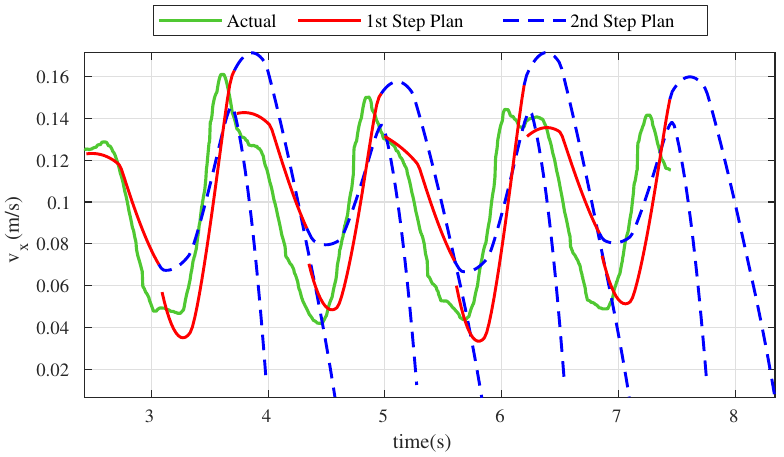}
\caption{Base velocity on x axis versus time for a flat terrain simulation. Red and blue are the optimized base velocity trajectories, during the first and and second footsteps respectively. Green is the actual velocity. Red is tracked by the controller and blue is discarded when a new trajectory is available.}
\label{base_v}
\end{figure}

In real-time simulations and real-world experiments, the planner computes base and feet trajectories for the next two steps using the latest state estimation, at every contact event. Then, the two step plan with timestamp is sent to the 400Hz TS-PIDC controller that runs on a second on-board computer. In our experiments only the first step from every planned trajectory is tracked by the controller before receiving a new input from the planner.  Although the second step of the plan is not used at the controller level, it plays a fundamental role in the trajectory optimization by acting as a statically stable attractor, which provides a zero velocity terminal cost without requiring the robot to have a zero velocity at the end of each step. This can be seen in Figure \ref{base_v}.

We tested the planner and controller on several cases in simulation other than flat terrains, namely; relocated box climbing, stair climbing (see video), gap crossing, obstacle avoiding footstep discovery and on the real robot, a relocated box (see video). We did not use any warm start, but instead we used quadratic costs with nominal final state references to guide the solution towards the desired tasks. The initial trajectory for linearization is merely the initial state estimation and static equilibrium inputs.

In simulation, stable base motions can be realised by penalizing the velocities and tangential directions of the contact forces. On real robot uneven terrain locomotion, additional intermediate position costs were added to move the base closer to support polygon. This makes the local optimum solution more reliant on the reference states, but they are needed to plan feasible trajectories due to model inaccuracies, tracking errors and occasional slips. Increasing the planning frequency is a possible solution, but the computation time remains a limiting factor.
\begin{figure}[t]
\centering
\includegraphics[width=0.9\linewidth]{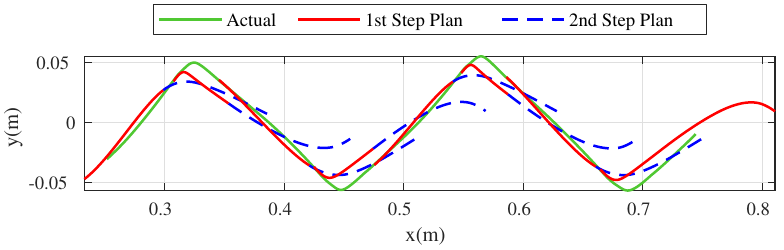}
\caption{Base position trajectories on the xy plane (i.e.\ overhead view) in a flat terrain simulation. Red and blue are the optimized base motions during the first and second steps, respectively. Green is the actual position. Red is tracked by the controller and blue is discarded when a new trajectory plan becomes available.}
\label{base_p}
\end{figure}

In Figure \ref{base_p} and \ref{base_v} the discontinuities between the consequent first plans are caused by the tracking errors and time spent during the trajectory optimization. The delay in the issuing a new plan implied that the TS-PIDC tracked the second step plan for a brief time before receiving the new step plan. 
The foot swing height is set prior to a task. For uneven terrain tasks, higher swing foot heights are used to be conservative against collisions with the environment. Also for uneven terrain, terms such as minimum clearance to a the edges and maximum distance between the feet are introduced. Although the analytical reachability cost itself encourages the latter, we have observed the benefits of these additional heuristics.

\begin{figure}[!t]
\centering
\includegraphics[width=0.9\linewidth]{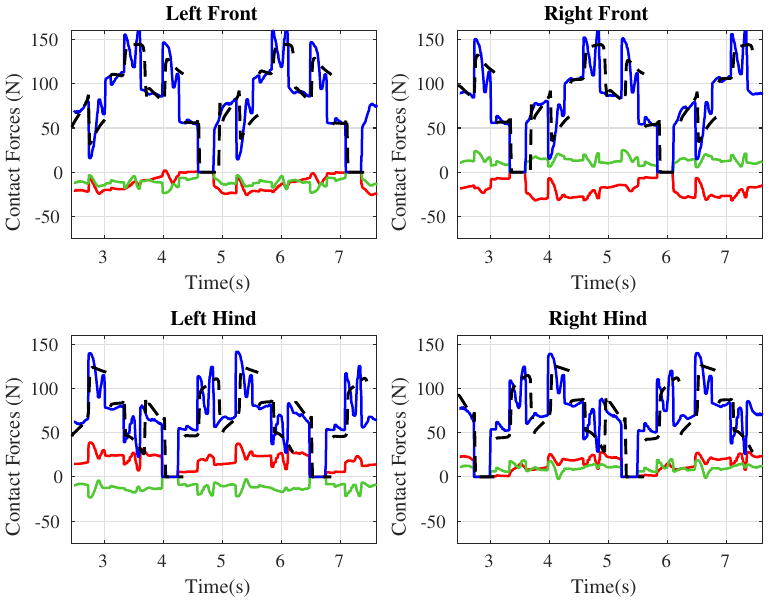}
\caption{Actual contact forces in flat terrain simulation on x axes (red), y axis (green) and z axis (blue) and planned contact force on z axis (black). Zero contact forces belong to the swing phases.}
\label{contact_lf_z}
\end{figure}

Contact forces are subject to high frequency changes, thus, contact forces from the trajectory optimisation are not tracked by the low-level controller. The TS-PIDC optimizes the contact forces via a constrained QP. In flat terrains, contact forces in z direction have a unique solution for a base state in the reduced momentum model. For this reason, simulation contact forces are similar to the trajectory optimization contact forces in z direction as seen in Fig. \ref{contact_lf_z}.

The box in both the simulation and real world experiment demonstrated is 15cm in height. In this scenario, the box is relocated to an arbitrary position on the movement direction of the robot at an arbitrary time to test the online optimization capability of the robot. As demonstrated in the simulation, the planner can generate a new feasible trajectory even when the previous trajectory would violate the new contact surface constraints.

In the real world experiment, a VICON Tracker system is used for the global localization of the robot base and the movable box. To demonstrate the effectiveness of online replanning, the box is moved in front of the robot during locomotion. When and where the box would be moved is not predetermined.

\begin{figure}[ht]
\centering
\includegraphics[width=0.9\linewidth]{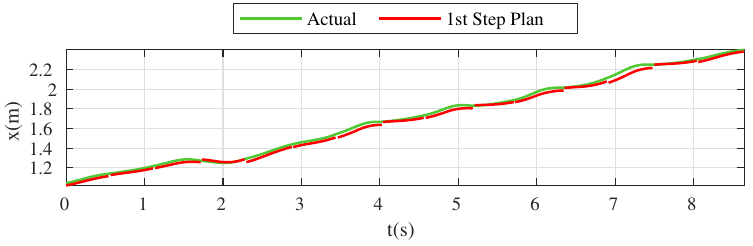}
\includegraphics[width=0.9\linewidth]{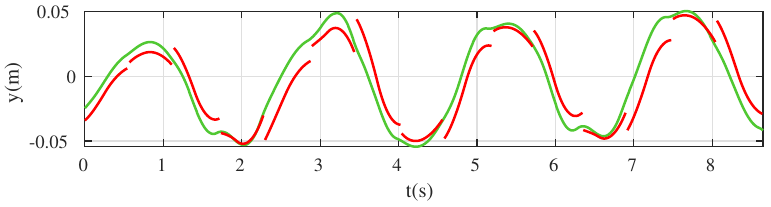}
\includegraphics[width=0.9\linewidth]{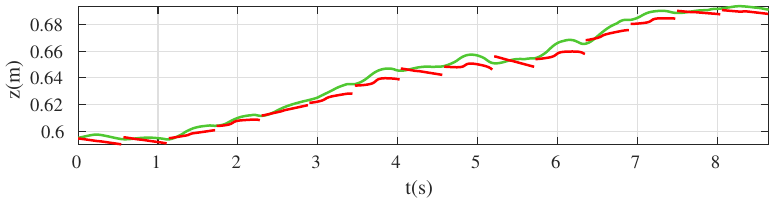}
\includegraphics[width=0.9\linewidth]{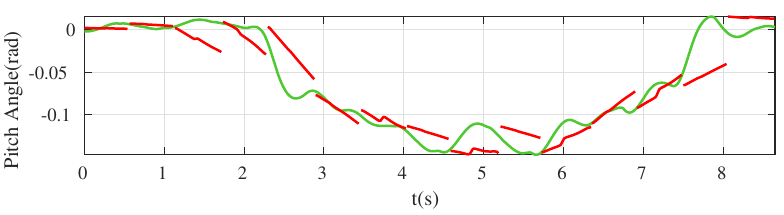}
\caption{Base position x, y, z and pitch angle (Euler Y) trajectories and state estimation of gap crossing simulation in real time. The figures cover 14 steps (3.5 gait cycles) and start with four feet on the first box, end with four feet on the second box.}
\label{base_p_gap}
\end{figure}

The gap crossing scenario shown in Figure \ref{fig:snap_sim} consists of two planes 30cm apart where the second plane is 10cm higher. Our method treats the gap as a rectangular box volume not to be assigned any contact on, otherwise it is treated as a climbing down and up task. When a nominal footstep is assigned on a gap, it is checked if it covers more than half of the gap. If it can, then the footstep is shifted forward to the second plane, otherwise it is shifted backwards to the first plane. Base position and pitch angle orientation states for the 14 footstep gap crossing with height difference task are shown in Figure \ref{base_p_gap}.

\begin{figure}[ht]
\centering
\includegraphics[width=0.8\linewidth]{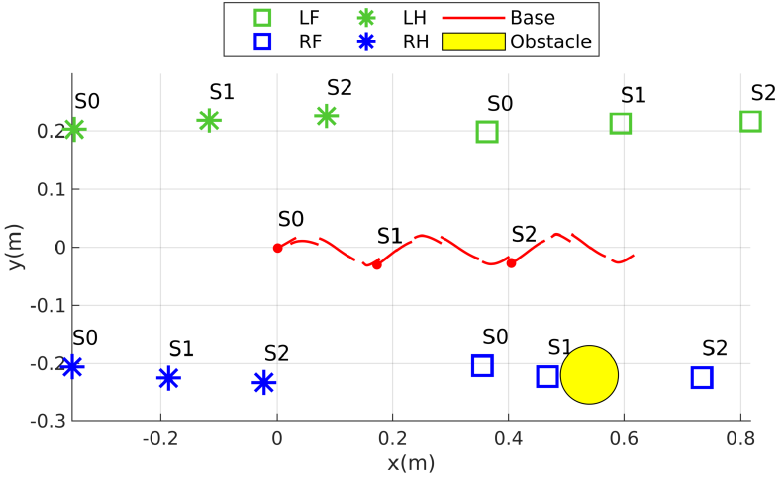}
\caption{Base and footstep planned trajectories of a obstacle avoiding footstep discovery task. $s_n$ indicates the base and feet position at the start of the i-th gait cycle. The base start points are indicated with red dots. The obstacle is a 10 cm diameter sphere (yellow).}
\label{base_feet_p}
\end{figure}

In footstep position discovery and obstacle avoidance scenario, we defined a base task and let the optimizer converge to footstep positions that are locally optimal to the base task on flat terrain by setting the $\bm{\mathrm{Q}}_{\mathrm{footstep}}$ lower than the $\bm{\mathrm{Q}}_{\mathrm{base}}$. We provided infeasible nominal footsteps, yet the optimizer converged to feasible footsteps that satisfy the base task. We then introduced a 10 cm diameter spherical obstacle as an inequality constraint and the resulting footsteps successfully avoided stepping on it. We used 4 step horizon in the footstep discovery and obstacle avoidance task. The base and footstep trajectories are shown in Fig. \ref{base_feet_p} and simulation is presented in the video.
\subsection{Limitations and Future Work}
In the test scenarios, four feet support phase duration is less than 400 ms and the solver needs to converge to a feasible solution within this time window. The main reason behind the choice of number of footstep in the horizon is that the convergence of a two footstep trajectory without warm-start takes roughly 100 ms on the on-board PC of the robot, depending on the gait phase duration and terrain constraints. Convergence for higher number of footstep require more iterations with increased computation time per iteration. 

In our method, the duration and equation of the switching constraints change with the moving horizon. Thus, the nonlinear optimal control problem is reinitialized each time the robot makes a new contact. Eliminating the computation cost of the reinitialization would reduce computation times and allow online optimization for larger number of footsteps.

Reachability is represented as an analytical cost function in the proposed optimization method. Although such a cost function requires less computation and is less sensitive to initial guess, it can perform poorly when a foot in the final phase is far from the 2D least squares (LS) fit plane of contacts. 
A nonconvex reachibility cost may perform better if provided with an adequate initial guess \cite{fankhauser2018robust}. 

The main limitation with the footstep discovery is that the contact planes are selected through the vertical projection of the nominal footsteps and/or heuristics. Introducing contact planes as decision variables would lead to a Mixed-Integer problem. Thus, the obstacle avoiding footstep position discovery is implemented for a convex terrain with a convex obstacle. A high level motion planner such as Humanoid Path Planner(HPP)\cite{tonneau2018efficient} can be used to generate the foothold and base sequence $E, \mathrm{X}$ for the Algorithm \ref{alg:NLOC}.

We provided minimal amount of reference state for the optimisation. Running cost state reference is time invariant and both running and final cost state references come from the nominal base position and footsteps. In this case, having a horizon of two or more steps is critical. Shorter horizons need more sophisticated, time variant reference states and providing these reference states itself becomes a trajectory optimization problem. 

Our framework has limited push recovery capacity due to optimizing only at feet touch down instants. Reduced computation time can allow intermediate trajectory optimizations during the swing phases and improve this aspect by generating reactive footstep motions similar to the capture point method \cite{pratt2006capture}. In fact, achieving uneven terrain locomotion and push recovery behavior within the same trajectory optimization framework remains an open problem.
\section{Conclusion}
In this paper, we propose a framework for online dynamic trajectory optimization and control for quadruped robots. The online trajectory optimization is based on optimizing base and footstep trajectories simultaneously around a nominal sequence. The nominal sequence both serves as a goal and a statically stable attractor for the constrained optimizer to find dynamically feasible trajectories. The trajectory optimizer runs at every foot touchdown for online replanning. A task space projected inverse dynamics controller tracks base and feet trajectories and may sacrifice tracking performance to satisfy physical constraints when necessary. The application of the framework on non-coplanar terrain is demonstrated in simulation and real world experiments. Future work will focus on increasing the computation efficiency to optimize for longer horizons online and enable intermediate optimizations during swing phase.




\section*{APPENDIX}

\subsection{QP form from LS}
\begin{figure}[thpb]
\centering
\includegraphics[width=0.5\linewidth]{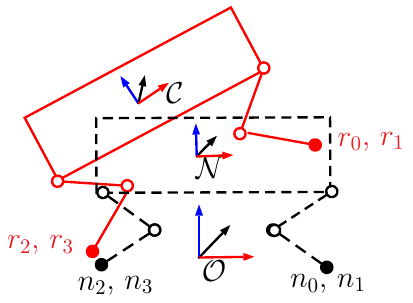}
\caption{Geometric representation of the reachability approximation. Nominal posture is dashed black and the altered posture solid red body. N and C are the body frames of the nominal and altered posture and O is the world frame.}
\label{appen_drawing}
\end{figure}

The reachability term is the result of the LS solution of the interpolation between a nominal posture and an altered posture, as shown in Figure \ref{appen_drawing}.
In Figure \ref{appen_drawing},  $h_{n}=\overrightarrow{\mathcal{ON}}_{z}$ and $h_{c}=\overrightarrow{\mathcal{OC}}_{z}$ are nominal and altered base height, $n_{i}$ are nominal feet positions, $r_{i}$ are feet position states in Eq. \ref{eq:ReducedMomentumModel}. $w_{x}=\overrightarrow{n_{2}n_{0}}_{x}=\overrightarrow{n_{3}n_{1}}_{x}$ and $w_{x}=\overrightarrow{n_{2}n_{3}}_{y}=\overrightarrow{n_{0}n_{1}}_{y}$ are nominal x, y distances between the feet.
\begin{align*}
    &\alpha_{x}=\mathrm{arcsin}(h_{c}/w_{n_{x}}),\quad \alpha_{y}=\mathrm{arcsin}(h_{c}/w_{n_{y}})\\
    &q_{x}=\mathrm{tan}(\alpha_{y})*h_{n}/h_{c}, \quad q_{y}=\mathrm{tan}(\alpha_{x})*h_{n}/h_{c}\\
    &r_{x}=\frac{1}{2} \alpha_{x} / h_{c}, \quad r_{y}=\frac{1}{2} \alpha_{y} / h_{c}
\end{align*}
Linear set of equations:
\begin{equation} \label{eq:linearequations}
    \begin{split}
    &p_{x}= \frac{1}{4}\sum_{i=0}^{3}r_{i_{x}}+\frac{1}{2}d_{x}(-r_{0}-r_{1}+r_{2}+r_{3})_{z}\\ 
    &p_{y}= \frac{1}{4}\sum_{i=0}^{3}r_{i_{y}}+\frac{1}{2}d_{y}(r_{0}-r_{1}+r_{2}-r_{3})_{z}\\ 
    &p_{z}=h_{n}+\frac{1}{4}\sum_{k=0}^{3}r_{i_{z}}\\
    &\theta_{x}=q_{x}(r_{0}-r_{1}+r_{2}-r_{3})_{z}\\
    &\theta_{y}=q_{y}(-r_{0}-r_{1}+r_{2}+r_{3})_{z}\\
    &\theta_{z}=0
    \end{split}
\end{equation}
Equation set (\ref{eq:linearequations}) is represented as $\bm{\mathrm{A}}\mathrm{x}=\bm{\mathrm{b}}$ and solved via LS $\frac{1}{2} \lvert \lvert \bm{\mathrm{A}}\mathrm{x}-\bm{\mathrm{b}} \rvert \rvert $. The intermediate quadratic cost function for the linear optimal control problem is in the form of
\begin{equation}
  \begin{split}
    L&=(\mathrm{x}-\mathrm{x}_{\mathrm{h}})^{T}\bm{\mathrm{Q}}_{\mathrm{h}}(\mathrm{x}-\mathrm{x}_{\mathrm{h}})\\
    &=\mathrm{x}^{T}\bm{\mathrm{Q}}_{\mathrm{h}} \mathrm{x}-2\mathrm{x}^{T}\bm{\mathrm{Q}}_{\mathrm{h}}\mathrm{x}_{\mathrm{h}}\\
    \end{split}  
\end{equation}
We write our LS problem in quadratic cost function form
\begin{equation}
  \begin{split}
    &\bm{\mathrm{Q}}_{\mathrm{h}}=\bm{\mathrm{A}}^{T}\bm{\mathrm{A}},\quad\mathrm{x}_{\mathrm{h}}=\bm{\mathrm{Q}}_{\mathrm{h}}^{+}\bm{\mathrm{A}}^{T}\bm{\mathrm{b}}\\
    \end{split}  
\end{equation}

\bibliographystyle{IEEEtran}
\balance
\bibliography{IEEEabrv,main}

\end{document}